\pdfoutput=1
\documentclass{new_tlp}

\usepackage{graphicx}
\usepackage{mathptmx}
\usepackage{tikz-qtree}
\usepackage{pgfplots}
\usepackage{filecontents}
\usepackage{csvsimple}
\usepackage{amsfonts,amsmath}
\usepackage{url}
\usepackage{hyperref}
\usepackage{verbatim}
\usepackage{color}

\definecolor{lgray}{gray}{0.95}
\definecolor{lblue}{rgb}{0.90,0.90,1.00}
\definecolor{lyellow}{rgb}{1.00,1.00,0.70}

\usepackage{listings}
\newtheorem{ex}{Example}

\newenvironment{codex}{\small\verbatim}{\endverbatim\normalsize}

\lstnewenvironment{code}
    {\lstset{}%
      \csname lst@SetFirstLabel\endcsname}
    {\csname lst@SaveFirstLabel\endcsname}
\newcommand{\BI}[0]{\begin{itemize}}
\newcommand{\EI}[0]{\end{itemize}}
\newcommand{\I}[0]{\item}
\newcommand{\BE}[0]{\begin{enumerate}}
\newcommand{\EE}[0]{\end{enumerate}}

\newcommand{\BX}[0]{\begin{ex}}
\newcommand{\EX}[0]{\end{ex}}
\newcommand{\BV}[0]{\begin{verbatim}}
\newcommand{\EV}[0]{\end{verbatim}}

\newcommand{\BF}[0]{\begin{filecontents*}{data.csv}}

\newcommand{\BQ}[0]{\color{blue}\begin{quote}}
\newcommand{\EQ}[0]{\end{quote}\color{black}}

\def \bscale1 {0.25}
\def \bscale {0.25}

\newcommand{\FIG}[4]{
\begin{figure}[htbp]
\centering
{\includegraphics[scale=#3]{figs/#4}}
\caption{#2}
\label{#1}
\end{figure}
}

\newcommand{\DFIG}[4]{
\begin{figure*}[htbp]
\centering
{\includegraphics[scale=#3]{figs/#4}}
\caption{#2}
\label{#1}
\end{figure*}
}

\begin{document}

\title{Interactive Text Graph Mining with a Prolog-based Dialog Engine}

\author[Paul Tarau and Eduardo Blanco]
         {
          Paul Tarau\\
          Dept. of Computer Science and Engineering \\ University of North Texas\\
          1155 Union Circle, Denton, Texas 76203, USA\\
           {\em paul.tarau@unt.edu}\\        
          \and Eduardo Blanco\\
          Dept. of Computer Science and Engineering \\ University of North Texas\\
          1155 Union Circle, Denton, Texas 76203, USA\\
           {eduardo.blanco@unt.edu}
          }

\date{}
\maketitle

\begin{abstract}

On top of a neural network-based dependency parser and a graph-based natural language processing  module we design a Prolog-based dialog engine that explores interactively a ranked fact database extracted from a text document.

We reorganize dependency graphs to focus on the most relevant content elements of a sentence and integrate sentence identifiers as graph nodes.
Additionally, after ranking the graph we take advantage of the implicit semantic information that dependency links  and WordNet bring in the form of subject-verb-object,
``is-a'' and ``part-of'' relations.

Working on the Prolog facts and their inferred consequences, the dialog engine specializes the text graph with respect to a query and reveals interactively the document's most relevant content elements.

The open-source code of the integrated system is available at \url{https://github.com/ptarau/DeepRank}.

\end{abstract}

\section{Introduction}\label{intro}

This is an extended and improved version of 
our PADL'20 paper \cite{padl20}.\footnote{
Selected by the reviewers of PADL'20 and the program chairs
Ekaterina Komendantskaya and Yanhong Annie Liu to be
submitted to the  Rapid Publications track of the
journal \emph{Theory and Practice of Logic Programming}.
}

Logic programming languages have been used successfully for inference and planning in natural language processing tasks restricted to narrow domains
\cite{actlan17,inclezan18,baral19,inclezan19}.
Their success, however, is limited in open-domain large-scale information extraction and knowledge representation tasks.
On the other hand, deep learning systems are good at basic tasks
ranging from parsing to factoid question answering,
but they are still taking baby steps emulating human-level inference
on complex documents \cite{bert18,bert19}.
Thus, a significant gap persists between neural and symbolic approaches in the field.


The work presented here aims at filling this gap.
We explore synergies between neural, graph-based and symbolic approaches
to solve a practical problem:
building a dialog agent.
This agent digests a text document (e.g., a story, a textbook, a scientific paper, a legal document)
and enables the user to interact with the most relevant content.


We will start with a quick overview of the system, including the main tools and techniques of each module.
Our system builds upon state-of-the-art natural language processing tools,
and
couples them with a declarative language module focusing on high-level text mining.
We integrate the modules in the Python-based nltk ecosystem \cite{bird-loper-2004-nltk},
and rely on the Java-based Stanford CoreNLP toolkit \cite{coreNLP} for basic
natural language processing tasks such as
sentence boundary detection, tokenization, part-of-speech tagging and parsing.


\subsubsection*{Overview of the System Architecture}

Fig. \ref{deepsys} summarizes the architecture of our system. The Stanford CoreNLP dependency parser is started as a separate server process to which the Python-based text processing module connects as a client. It interfaces with the Prolog-based dialog engine by generating a clausal representation of the document's structure and content as well as the user's queries. The dialog engine is responsible for handling the user's queries for which answers are sent back to the Python front-end,
which also handles calls to OS-level spoken-language services, when activated.

\FIG{deepsys}{System Architecture}{0.42}{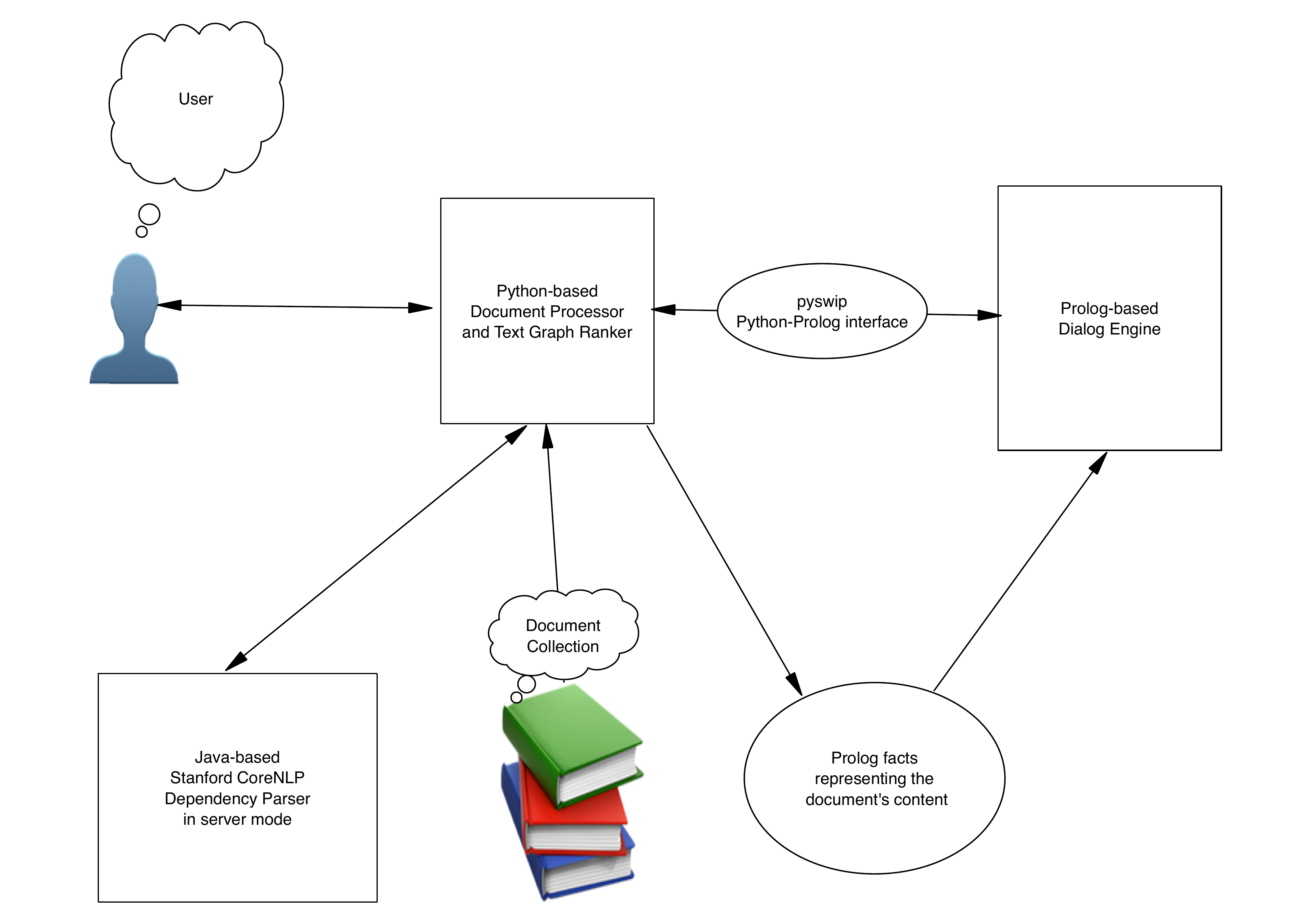}

State-of-the-art dependency parsers \cite{stan,AdolphsXLU11,choi:17a},
among which the neural Stanford dependency parser \cite{stan} stands out, 
produce highly accurate dependency graphs.
The vertices in these graphs are words and their part-of-speech tags,
and labeled edges indicate the syntactic heads of words (e.g., subject, direct object).
In contrast to collocations in a sliding window,
dependency graphs provide ``distilled'' building blocks through which
 a graph-based natural language processing system can absorb
 higher level linguistic information.


Inspired by the effectiveness of algorithms like Google's PageRank, recursive ranking algorithms applied to text graphs have enabled extraction of keyphrases, summaries and relations.
Their popularity continues to increase due to their holistic view on the interconnections between text units,
which signal the most relevant text units.
Additionally, these algorithms are comparatively simpler.
At more that 3100 citations and a follow-up of other highly cited papers \cite{Erkan:2004},
 the TextRank algorithm \cite{EMNLP:TR,ijcnlp05} and its creative descendants have  extended their  applications to a wide variety of document types and social media interactions in a few dozen languages. 
 

While part of the family of the TextRank descendants, our graph-based text processing algorithm  will use information derived from the dependency graphs associated to sentences.
We leverage part-of-speech tags assigned to vertices (words)
and edge labels (syntactic dependencies between words)
in dependency graphs
in order to extract rank-ordered facts corresponding to  content elements present in sentences.  We pass these to logic programs that can query them and  infer new relations, beyond those that can be mined directly from the text.

Like in the case of a good search engine,  interactions with a text document will focus on the most relevant 
and semantically coherent elements matching a query. 
With this in mind, the natural feel of an answer syntactically
appropriate for a query is less important than the usefulness of the content elements extracted: just sentences of the document in their natural order.

We will also enable spoken interaction with the dialog engine, opening the door to the use of the system via voice-based appliances. Applications range from assistive technologies to visually challenged people, live user manuals, teaching from K-12 to graduate level classes, and interactive information retrieval from complex technical or legal documents.

The paper is organized as follows.
Section \ref{tg} describes the graph-based Natural Language Processing module.
Section \ref{dia} describes our Prolog-based dialog engine.
Section \ref{inter} shows interaction examples with several document types.
Section \ref{disc} puts in context the main ideas of the paper and justifies some of the architecture choices we have made.
Section \ref{rel} overviews related work and background information.
Section \ref{conc} concludes the paper.

\section{The  graph-based Natural Language Processing module}\label{tg}
We have organized our Python-based textgraph processing algorithm together with the Prolog-based dialog engine
into a  unified system.\footnote{Our implementation is available at \url{https://github.com/ptarau/DeepRank.}}
We start with the building and the ranking of the text graph. Then, we overview the summary, keyphrase and relation extraction components,  and the creation of the Prolog database that constitutes the logical model of the document,
to be processed by the dialog engine.

\subsection{Building and ranking the text graph}

We connect as a Python client to the Stanford CoreNLP server and use it to provide
our dependency links via the wrapper at \url{https://www.nltk.org/} of the Stanford CoreNLP toolkit \cite{coreNLP}. 

Unlike the original TextRank and related approaches that develop special techniques for each text processing task, we  design a unified algorithm to obtain graph representations of documents, that are suitable for keyphrase extraction, summarization and interactive content exploration.

We use unique sentence identifiers and unique lemmas\footnote{A lemma is a canonical representation of a word, as it stands in a dictionary, for all its inflections e.g., it is `{\bf `be''} for ``is'', ``are'', ``was'' etc.}
as nodes of the text graph. As  keyphrases are centered around nouns and good summary sentences are likely to talk about important concepts,  we will need to reverse some links in the dependency graph provided by the parser, to prioritize nouns
and deprioritize verbs, especially auxiliary and modal ones. 
Thus, we (a) redirect the dependency edges toward nouns with subject and object roles, as shown for a simple short sentence  in Fig. \ref{depgraph},
 and (b) add
{\em ``about''} edges from the sentences they occur in.
 
\DFIG{depgraph}{Dependency graph of a simple sentence with redirected  and  newly added arrows }{0.38}{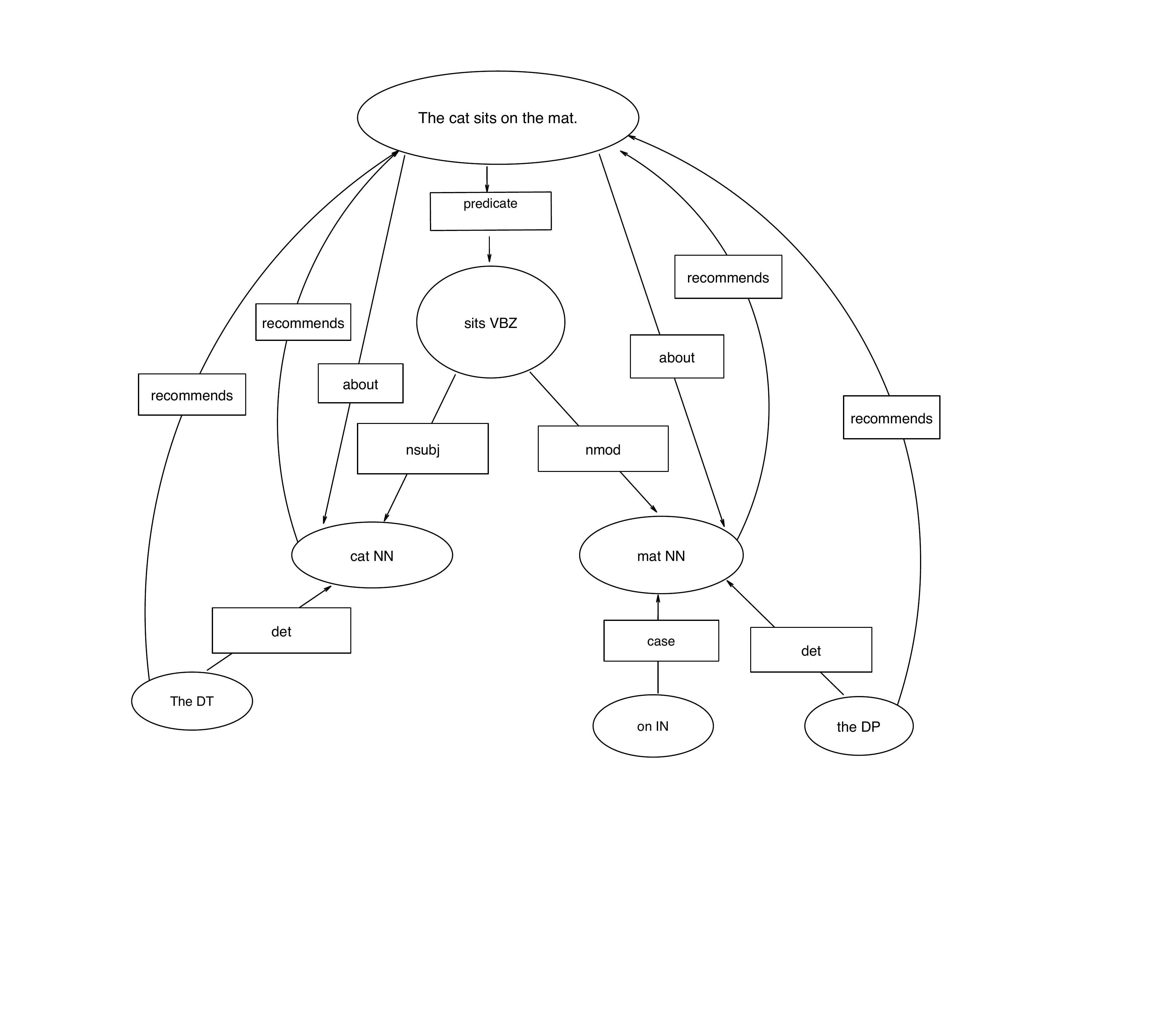}

We also create {\em ``recommend''} links  from words to the sentence identifiers and back from sentences to verbs with {\em predicate} roles
to indirectly ensure that sentences recommend and are 
recommended by their content. Specifically, we ensure that (a) sentences 
recommend verbs with predicate function,
and (b) their recommendation spreads to nouns that are
predicate arguments (e.g., having subject or object roles).
  
By using the PageRank implementation of the {\bf networkx} 
toolkit,\footnote{\url{https://networkx.github.io/}}
after ranking the sentence and word nodes of the text graph,
the system is also able to display
subgraphs filtered to contain only the highest ranked nodes, using Python's 
 {\tt graphviz} library.
 
An example of text graph, filtered to only show word-to-word links, derived from the U.S. Constitution,\footnote{Available as a text document at: \url{https://www.usconstitution.net/const.txt}}
is shown in Fig. \ref{constit}.
\FIG{constit}{Text graph fragment connecting the highest ranked words in the U.S. Constitution}{0.52}{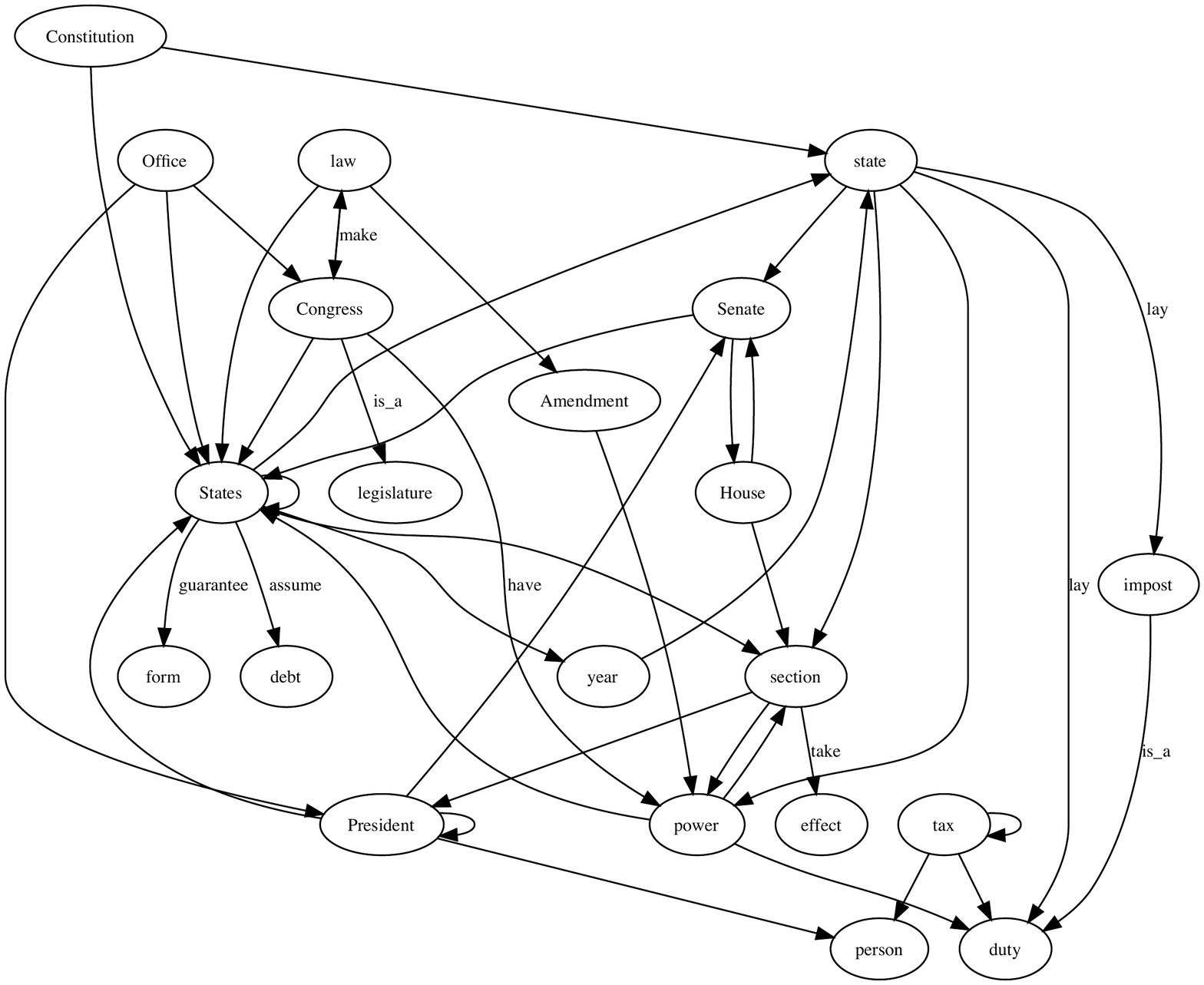}  
  
\subsection{Pre- and post-ranking graph refinements}

The algorithm induces a form of automatic stop word filtering, 
due to the fact that our dependency link arrangement ensures that modifiers with lesser semantic value relinquish their rank by pointing to more significant lexical components.
This is a valid  alternative to explicit ``leaf trimming''  before ranking, which remains an option for reducing graph size for large texts or multi-document collections as well as helping with a more focused relation extraction from the reduced graphs.

Besides word-to-word links, our text graphs connect sentences as additional dependency graph nodes, resulting in a unified keyphrase and summary extraction framework. Note also that, as an option that is relevant
especially for scientific, medical or legal documents, we add {\tt first\_in} links from a word 
to the sentence containing its first occurrence, to prioritize sentences where concepts 
are likely to be defined or explained.

Our reliance on graphs provided by dependency parsers builds a bridge between deep neural network-based machine learning and graph-based natural language processing enabling us to often capture  implicit semantic information.

\subsection{Summary and keyword extraction}

As link configurations  tend to favor very long sentences, a post-ranking normalization is applied for sentence ranking. 
After ordering sentences by rank we extract the highest ranked ones and reorder them in their natural order in the text to form a more coherent summary.

We use the parser's compound phrase tags to  fuse along dependency links. 
We   design our keyphrase synthesis algorithm to ensure that highly ranked  words will pull out their contexts from sentences, to make up meaningful keyphrases. As a heuristic, we mine for a context of 2-4 dependency linked words of a highly ranked noun, while ensuring that the context itself has a high-enough rank, as we compute a weighted average favoring the noun
over the elements of its context.

\subsection{Relation extraction}\label{rels}

We  add subject-verb-object facts extracted from the highest ranked dependency links, enhanced with ``is-a'' and ``part-of'' relations using WordNet via the {\tt nltk} toolkit. We plan in the future to also generate relations from conditional statements  identified following dependency links and involving negations, modalities, conjuncts and disjuncts, to be represented as Prolog rules. 
Subject-verb-object (SVO) relations are extracted directly from the dependency graph and an extra argument is added to the triplet marking the number of the sentence they originate from.

``{Is-a}''  relations are extracted using WordNet \cite{Fellbaum98} hypernyms 
and 
hyponyms.\footnote{More general and, respectively, more specific concepts.}
Similarly, ``{\tt part\_of}'' relations are extracted using 
meronyms and 
holonyms.\footnote{Concepts corresponding to objects that are part of, and, respectively,  have as part other objects.}
As a heuristic that ensures that they are relevant to the content of the text, we ensure that both their arguments are words that occur in the document, when connecting their corresponding synsets via WordNet relations. By constraining the two ends of an ``is-a'' or ``part-of'' edge to occur in the document, we avoid relations derived from  synsets unrelated to the document's content. In fact, this provides an  effective word-sense disambiguation heuristic.

\section{The Prolog-based dialog engine}\label{dia}

After our Python-based document processor, with help from the Stanford dependency parser, builds and ranks the text graph and extracts summaries, keyphrases and relations, we pass them to the Prolog-based  dialog engine.

\subsection{Generating  input for post-processing by logic programs}

Once the document is processed, we generate,
besides the dependency links provided by the parser,
relations containing  facts that we have gleaned 
from processing the document. Together, they form
a Prolog database representing the content of the document.

To keep the interface simple and portable to other logic programming tools, we generate the following predicates in the form of Prolog-readable code, in one file per document:
\BE
\I {\tt keyword(WordPhrase).} --  the extracted keyphrases
\I {\tt summary(SentenceId,SentenceWords).} --  the extracted summary sentence identifiers and list of words in sentence
\I {\tt  dep(SentenceID,WordFrom,FromTag,Label,WordTo,ToTag).} -- a component of a dependency link, with the first argument indicating the sentence they have been extracted
\I {\tt  edge(SentenceID,FromLemma,FromTag,RelationLabel,ToLemma,ToTag).} --  edge marked with sentence identifiers indicating  where it was extracted from, and the lemmas with their POS tags at the two ends of the edge

\I {\tt rank(LemmaOrSentenceId,Rank).} -- the rank computed for each lemma
\I {\tt w2l(Word,Lemma,Tag).} -- a map associating to each word a lemma and a tag, as found by the POS tagger
\I {\tt svo(Subject,Verb,Object,SentenceId).} -- subject-verb-object relations extracted from parser input or WordNet-based {\tt is\_a} and {\tt part\_of} labels in verb position 
\I {\tt ner(SentId,ListOfNamedEntityPairs)} extracted from Named Entity Recognizer ({\tt NER}) annotations
\I {\tt sent(SentenceId,ListOfWords).} -- the list of sentences in the document with a sentence identifier as first argument and a list of words as second argument
\EE
These predicates provide a relational view of a document in the form of a fact database that  will support  the inference mechanisms built on top of it.

The resulting logic program can then be processed  with Prolog semantics, possibly enhanced by using 
constraint solvers \cite{OzEngines:97}, 
abductive reasoners  \cite{abduct02}
or via Answer Set Programming systems \cite{ijcai2018-769,asp}. Specifically, we expect benefits from such extensions for tackling computationally difficult problems like word-sense disambiguation (WSD) or entailment inference
as well as domain-specific reasoning \cite{inclezan19,naraction,baral19}.

We have applied this process to the {\em Krapivin document set} \cite{dataset08}, a  collection of  {\bf 2304} research papers annotated with the authors' own keyphrases and abstracts.
  
The resulting 3.5 GB {\em Prolog 
dataset\footnote{\url{http://www.cse.unt.edu/~tarau/datasets/PrologDeepRankDataset.zip}}} 
is made available for researchers in the field, interested to explore declarative reasoning or text mining mechanisms. 

\subsection{The Prolog interface}

We use as a logic processing tool the open 
source SWI-Prolog  system\footnote{\url{http://www.swi-prolog.org/}}
 \cite{swi} that
 can be called from, and can call Python programs using
 the {\tt pyswip} adaptor.\footnote{\url{https://github.com/yuce/pyswip}}
 After the adaptor creates the Prolog process and the content of the 
 digested document
 is transferred from Python (in a few seconds for typical 
 scientific papers with 10-15 pages), query processing is realtime.

\subsection{The user interaction loop}

With the Prolog representation of the digested document in memory, the dialog starts by displaying the summary and keyphrases extracted from the 
document.\footnote{And also speak them out if the {\tt quiet} flag is off.}
One can see this as a ``mini search-engine'', specialized to the document, and, with help of an indexing layer, extensible to multi-document collections. The dialog agent associated to the document answers queries as sets of salient sentences extracted from the text, via a specialization of our summarization algorithm to the context inferred from the query.

As part of an interactive  {\em read/listen, evaluate, print/say} loop,
we generate for each query sentence, a set of predicates that
are passed to the Prolog process, from where answers will come back
via the {\tt pyswip} interface.
The predicates extracted from a query have the same structure as the 
database 
representing the content of the complete document,
initially sent to Prolog.

\subsection{The answer generation algorithm}

Answers are generated by selecting the most relevant sentences, presented
in their natural order in the text, in the form of a specialized ``mini-summary''. We will next overview our query answering algorithm, with examples
to follow in section \ref{inter}.

\subsubsection{Query expansion}

Answer generation starts with a query-expansion mechanism
via relations that are derived by
finding, for lemmas in the query,
WordNet hypernyms, hyponyms, meronyms and holonyms,
as well as by directly extracting them 
from the query's dependency links.
We use the rankings available both in the query
and the document graph to prioritize the highest ranked sentences connected to the highest ranked  nodes in the query.

\subsubsection{Short-term dialog memory}

We keep  representations of recent queries in memory, as well as the
answers generated for them.
If the representation of the  current query overlaps with a past one, we use content in the past query's database to extend query expansion
to cover edges originating from that query.
Overlapping is detected via
shared edges between noun or verb nodes 
between the query graphs.


\subsubsection{Answer sentence selection}

Answer sentence selection is performed with a combination of several
interoperating algorithms:
\BI
\I use of {\em personalized PageRank} \cite{perso1,perso2} with a dictionary provided by highest ranking lemmas and their ranks in the query's graph, followed by reranking the document's graph to specialize to the query's content
\I matching guided by SVO-relations
\I matching of edges in the query graph against edges in the document graph
\I query expansion guided by rankings in both the query graph and the document graph
\I matching guided by a selection of related content components in the short-term dialog memory window
\EI
 
Matching against the Prolog database representing the document
is  implemented  as a size constraint on the 
intersection of the expanded
query lemma set, built with highly ranked shared lemmas pointing to sentences
containing them.
The set of answers is organized to return the highest-ranked sentences based 
on relevance to the query and in the order in which they appear in the document.

We keep the dialog window relatively small (limited to the highest ranked 3 sentences in the answer set, by default). 
Relevance is ensured with help from the rankings computed for both
the  document content and the query. 

\subsubsection{Personalized PageRank}

Using {\em personalized PageRank} \cite{perso1,perso2} can be seen as a specialization of the document graph with respect to the query. The personalization dictionary is also used to implicitly redirect flow, otherwise stuck in the sink nodes of the text graph, to
 content related to the query. The predicate {\tt query\_pers\_sents} is generated for each query and then passed to Prolog, where it is used to prioritize the answers computed by the answer search algorithms.

\subsubsection{Matching guided by SVO-relations}

SVO-facts  inferred from syntactic dependencies and from WordNet relations have a 4-th argument, indicating the sentence number where they have been found. Thus walking over them in a transitive closure computation (limited to at most K inference steps) allows us to collect
a path made of the sentences and relations used at each step, suggesting 
possibly interesting candidate answers.

The predicate {\tt tc/7} implements a {\tt K}-step limited transitive closure computation
also controlled by a set of relations available
and a loop checking mechanism that avoids revisiting the same nodes repeatedly.
We believe that it also shows the expressiveness of a declarative programming pattern
in handling a fairly complex set of requirements in a clear and compact form.
Our computation is exposed via the interface predicate 
{\tt tc/5}. 

{\em The predicate {\tt tc(K,A,Rels,C,Res)}  holds 
if we can get from word {\tt A} to word {\tt C}
in at most {\tt K} steps
using any relation in the set {\tt Rels} and returning
in {\tt Res} the number of steps left, the path followed and
a sentence number in the document, possibly relevant as an answer.}
\begin{code}
%% tc(+K,+A,+Rels,?A,-Res)
tc(K,A,Rels,C,Res):-tc(A,Rels,C,[],K,_,Res).
\end{code}

\begin{code}
tc(A,Rels,C,Xs,SN1,N2,Res) :-
  succ(N1,SN1),
  member(Rel,Rels),
  call_svo(A,Rel,B,Id),
  not(memberchk(B-_,Xs)),
  tc1(B,Rels,C,[A-Rel|Xs],Id,N1,N2,Res).

tc1(B,_Rels,B,Xs,Id,N,N,res(N,Id,Xs)):-
  % Id must be a known sentence Id!
  nonvar(Id).
tc1(B,Rels,C,Xs,_,N1,N2,Res) :-
   tc(B,Rels,C,Xs,N1,N2,Res).
\end{code}
Note that loop checking is achieved by keeping  a path of elements the form Word-Relation and
the the ``{\tt \_}'' variable in the definition of {\tt tc/5} ensures that paths of length {\em up to}
{\tt K} are returned.
We  also accommodate {\tt SVO} relations originating from a {\em domain-specific ontology},
for which the sentence identifier is left as an {\em unbound logical variable}, provided
that at the end of the available 
inference steps an actual sentence of the document
is returned, a requirement ensured with the {\tt nonvar/1} test 
in the first clause of the predicate {\tt tc1/8}.
If the system detects the presence of additional {\tt svo/4} facts from
a domain specific ontology (consulted using a plugin mechanism), the single step
predicate {\tt call\_svo/4} will extend its scope over 
such additional {\tt SVO} relations.

\subsubsection{Using Named Entity Recognition}

Named Entity Recognition (NER) provides person, location, time, etc. annotations that provide additional relations, usable for inference.
When NER is available, matching directed by {\bf wh}-words like {\em where}, {\em when}, {\em who}, questions trigger scanning the named entity database, also sent to Prolog.

For a document on the CDC COVID-19 status, the system extracts {\tt NER} relations like
\begin{code}
ner(86, [(0, ('March', 'DATE')), 
           (1, ('10', 'DATE')), 
           (2, ('CDC', 'ORGANIZATION')), 
           (3, ('infection', 'CAUSE_OF_DEATH'))]).
\end{code}
saying that  sentence {\tt 86} is talking about named entities
hinting at events on March 10 when the CDC, an organization,
has identified life-threatening infections.
 
\subsubsection{Using {\bf wh}-word replacement with logic variables}

Edges in the query originating or targeting {\bf wh}-words get
replaced with logic variables with labels also corresponding to
expected syntactic roles (e.g., {\tt nsubj} for {\rm who}).

Besides matching them against corresponding edges in the document
we also take advantage of the named entity relation {\tt ner/2}
from which we derive specialized answer predicates.
\begin{code}
who(KWs,SentId):-wh(['PERSON','ORGANIZATION','TITLE'],KWs,SentId).

where(KWs,SentId):-
  wh(['LOCATION','CITY','COUNTRY','STATE_OR_PROVINCE'],KWs,SentId).
  
many(KWs,SentId):-wh(['NUMBER', 'ORDINAL', 'MONEY'],KWs,SentId).

when(KWs,SentId):-wh(['DATE','TIME','DURATION'],KWs,SentId).
\end{code}
The {\tt wh/3} predicate will scan the corresponding {\tt ner/2}
facts against a match, to be also validated by ensuring that answers
have a high enough personalized PageRank.

\section{Interacting with the dialog engine}\label{inter}

We will next show interaction examples with several document types, with
focus on key aspects of our question-answering algorithms.

The following example shows the result of a query on the US Constitution document, with edges marked by
POS-tags of the two nodes aggregated with the label of the dependency
link  connecting them.

\FIG{remove}{Graph of a query on the U.S. Constitution}{0.44}{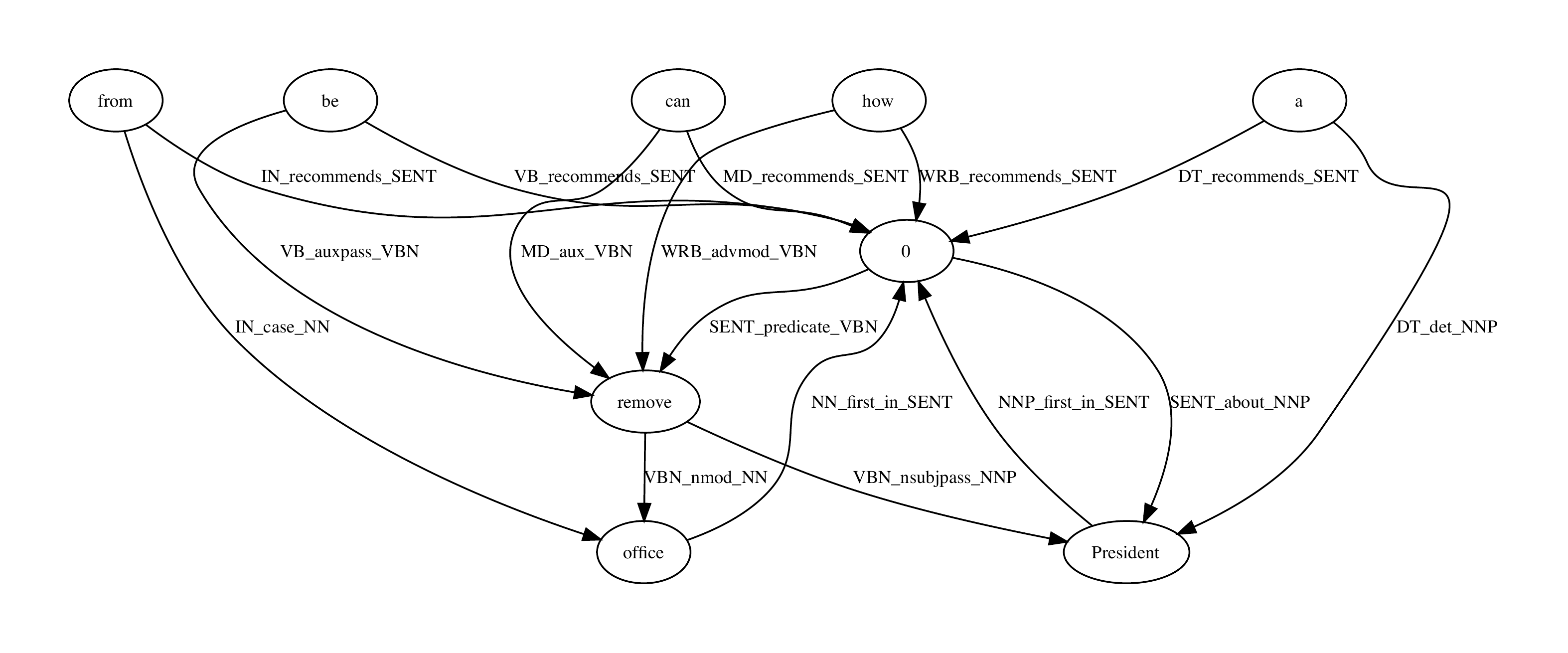}

{\small
\begin{quote}
{\verb~>>>~ talk\_about('examples/const')}

{\em
{\bf \verb~?--~ How can a President be removed from office?}

59 : In Case of the Removal of the President from Office , or of his Death , Resignation , or Inability to discharge the Powers and Duties of the said Office , the same shall devolve on the Vice President , and the Congress may by Law provide for the Case of Removal , Death , Resignation or Inability , both of the President and Vice President , declaring what Officer shall then act as President , and such Officer shall act accordingly , until the Disability be removed , or a President shall be elected . 

66 : Section 4 The President , Vice President and all civil Officers of the United States , shall be removed from Office on Impeachment for , and Conviction of , Treason , Bribery , or other high Crimes and Misdemeanors . 

190 : If the Congress , within twenty one days after receipt of the latter written declaration , or , if Congress is not in session , within twenty one days after Congress is required to assemble , determines by two thirds vote of both Houses that the President is unable to discharge the powers and duties of his office , the Vice President shall continue to discharge the same as Acting President ; otherwise , the President shall resume the powers and duties of his office .
}
\end{quote}
}

Note the relevance of the extracted sentences and resilience to semantic and syntactic variations (e.g., the last sentence does not contain the word ``remove'').
The dependency graph of the query is shown in Fig. \ref{remove}. The clauses of the {\tt query\_rank/2} predicate in the Prolog database corresponding to the query are:
\begin{codex}
query_rank('President', 0.2162991696472837).
query_rank('remove', 0.20105324712764877).
query_rank('office', 0.12690425831428373).
query_rank('how', 0.04908035060099132).
query_rank('can', 0.04908035060099132).
query_rank('a', 0.04908035060099132).
query_rank('be', 0.04908035060099132).
query_rank('from', 0.04908035060099132).
query_rank(0, 0.0023633884483800784).
\end{codex}

The impact of Personalized PageRank can be seen by comparing the cloud-map for the document (see Fig. \ref{dcloud}) with the cloud-map of the document specialized to the query (see Fig. \ref{qcloud}). Note the increased emphasis in Fig. \ref{qcloud} of some relevant actors (President, Senate) as well as the relevant concepts related to removal from office of a President (impeachment, profit, disability, judgment).

\FIG{dcloud}{Word-cloud of U.S. Constitution before specialization w.r.t. to query}{0.72}{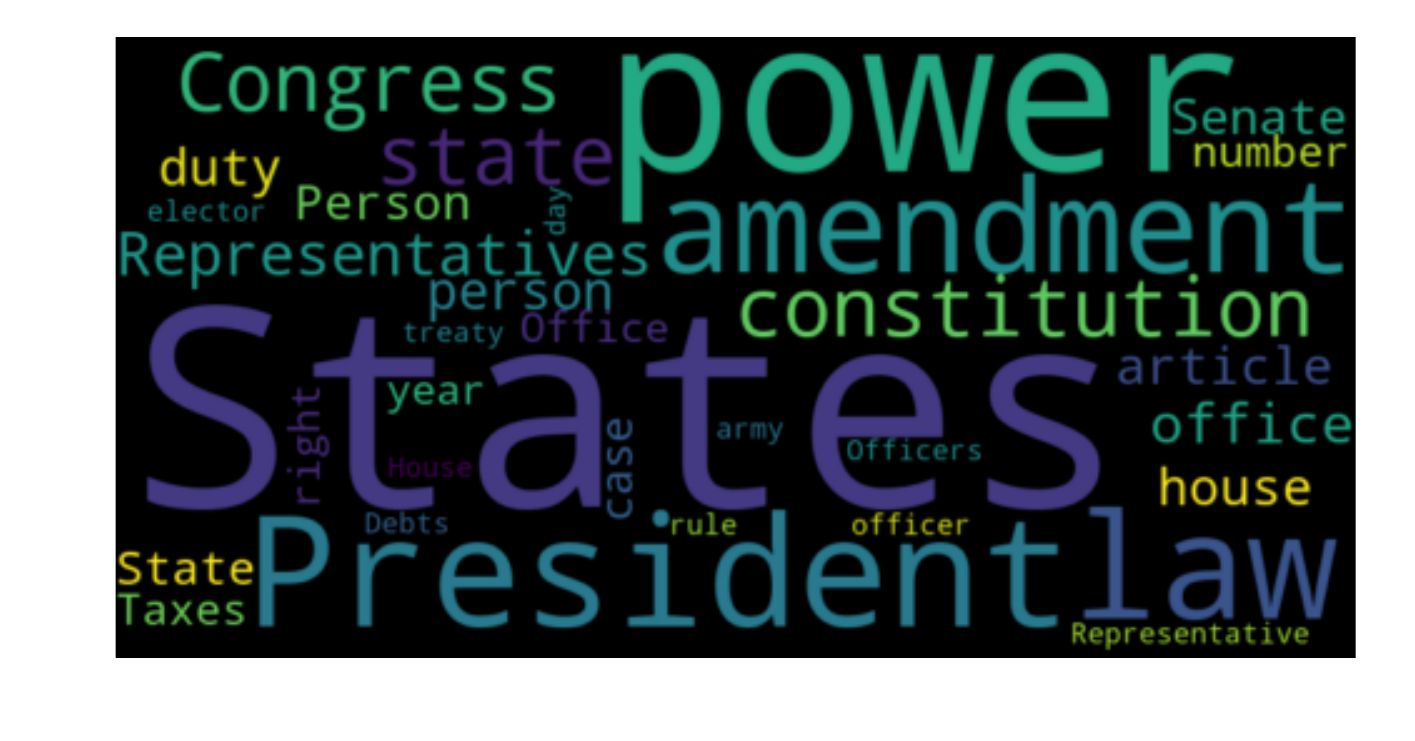} 
\FIG{qcloud}{Word-cloud of U.S. Constitution after query-driven personalized PageRank}{0.72}{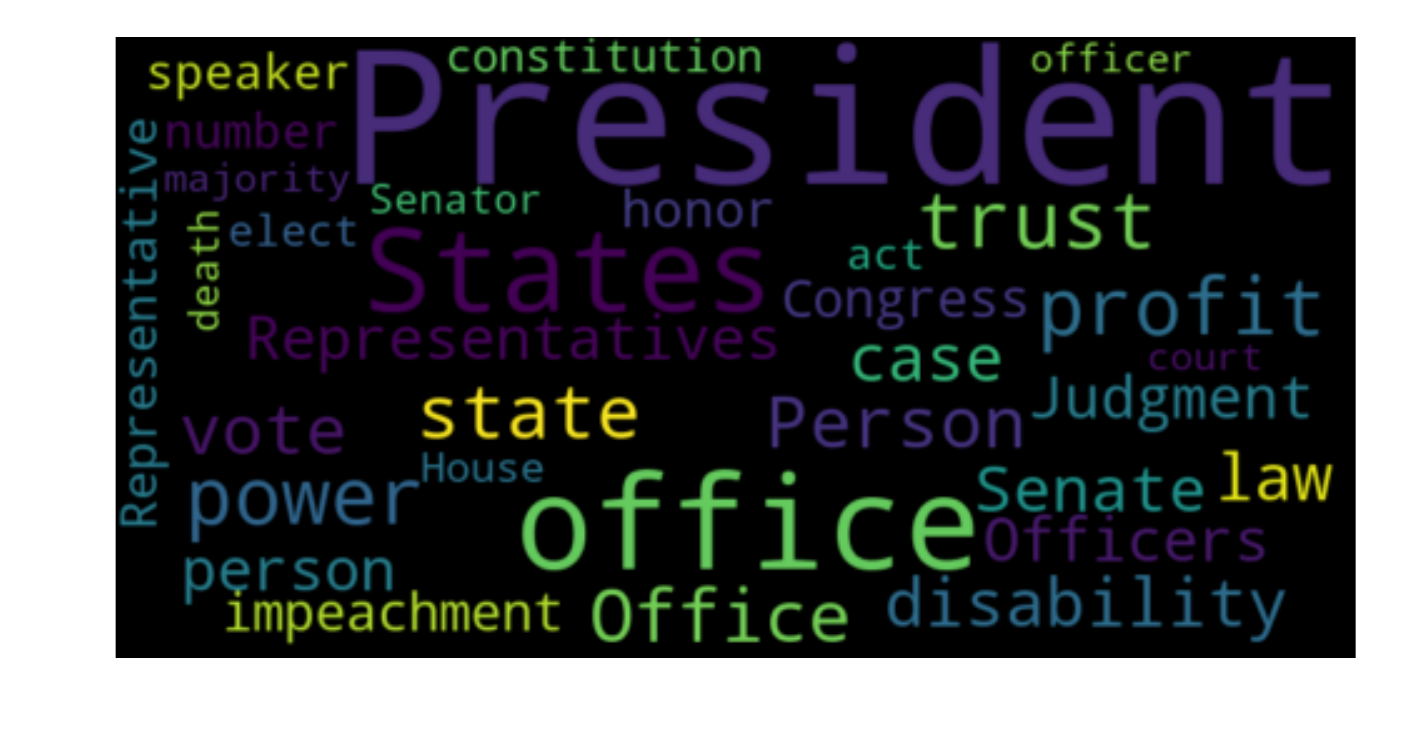}

Our next example uses an ASCII version of
Einstein's 1920 book on relativity, retrieved from the Gutenberg
collection\footnote{
{\small \url{https://www.gutenberg.org/files/30155/30155-0.txt}}
}
and trimmed to the actual content of the book (250 pages in {\tt epub} form).

{\small
\begin{quote}
{\verb~>>>~ talk\_about('examples/relativity')}

{\em
{\bf \verb~?--~ What happens to light in the presence of gravitational fields?}

611 : In the example of the transmission of light just dealt with , we have seen that the general theory of relativity enables us to derive theoretically the influence of a gravitational field on the course of natural processes , the laws of which are already known when a gravitational field is absent . 

764 : On the contrary , we arrived at the result that according to this latter theory the velocity of light must always depend on the co-ordinates when a gravitational field is present . 

765 : In connection with a specific illustration in Section XXIII , we found that the presence of a gravitational field invalidates the definition of the coordinates and the time , which led us to our objective in the special theory of relativity . 
}
\end{quote}
}
\FIG{light}{Graph of query on Einstein's book on Relativity}{0.33}{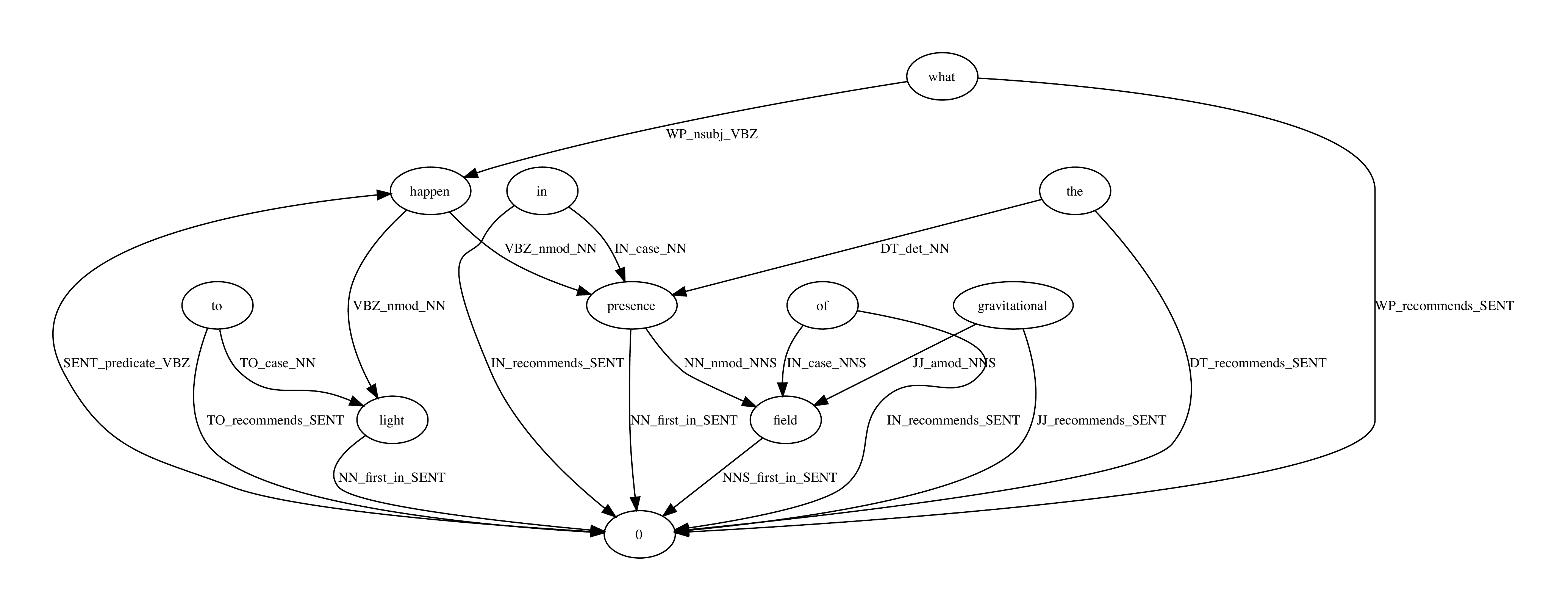}

The query graph is shown in Fig. \ref{light}.
After the less than 30 seconds that it takes to digest the book, answers
are generated in less than a second for all queries that we have tried.
Given the availability of spoken dialog, a user can iterate and refine
queries to extract the most relevant answer sentences of a document.

On an even larger document, like the Tesla Model 3 
owner's manual,\footnote{
\url{https://www.tesla.com/sites/default/files/model_3_owners_manual_north_america_en.pdf}
}
digesting the document takes about 60 seconds and results in 12 MB of Prolog clauses. After that, query answering is still below 1 second.
{\small
\begin{quote}
{\verb~>>>~ talk\_about('examples/tesla')}

{\verb~?--~ {\bf How may I have a flat tire repaired?}}

3207 : Arrange to have Model 3 transported to a Tesla Service Center , or to a nearby tire repair center . 

3291 : Note : If a tire has been replaced or repaired using a different tire sealant than the one available from Tesla , and a low tire pressure is detected , it is possible that the tire sensor has been damaged . 
\end{quote}
}
The highly relevant first answer is genuinely useful in this case, given that Tesla Model 3's do not have a spare tire. Being able to use voice queries while driving and in need of urgent technical information about one's car, hints towards obvious practical applications of our dialog engine.

Directly querying news articles with potentially urgent information is another application.
The following interaction is based on the US-CDC guidelines on the COVID-19 pandemics (as posted on March 17, 2020).

{
\begin{quote}
{\verb~>>>~ talk\_about('examples/covid')}

{\verb~?--~ {\bf How does the COVID-19 virus spread in a community?}}

25 : Pandemics happen when a new virus emerges to infect people and can spread between people sustainably . 

26 : Because there is little to no pre-existing immunity against the new virus , it spreads worldwide . 

47 : Three U.S. states are experiencing sustained community spread . 
 
\end{quote}
}

The highest ranked lemmas after running personalized PageRank show
the emphasis on key lemmas coming from the query as well as
some of the overall important related concepts in the document.
\begin{code}
query_pers_words('virus', 0.1279800580708069).
query_pers_words('spread', 0.08076646367741867).
query_pers_words('covid', 0.06794580306202914).
query_pers_words('community', 0.05973841781795769).
query_pers_words('CoV', 0.01846080235270702).
query_pers_words('response', 0.012416829623738527).
query_pers_words('influenza', 0.012240656425061098).
query_pers_words('culture', 0.012084585427826857).
query_pers_words('illness', 0.01129131793978254).
query_pers_words('people', 0.009611356400409315).
query_pers_words('healthcare', 0.008738409714154799).
query_pers_words('risk', 0.00859884680785955).
\end{code}

The evolution between the cloud-map of the document and the cloud-map after the reranked document is shown in Fig. \ref{dcovid} and Fig. \ref {qcovid}.

\FIG{dcovid}{Word-cloud of the CDC COVID-19 document before specialization w.r.t. to query}{0.72}{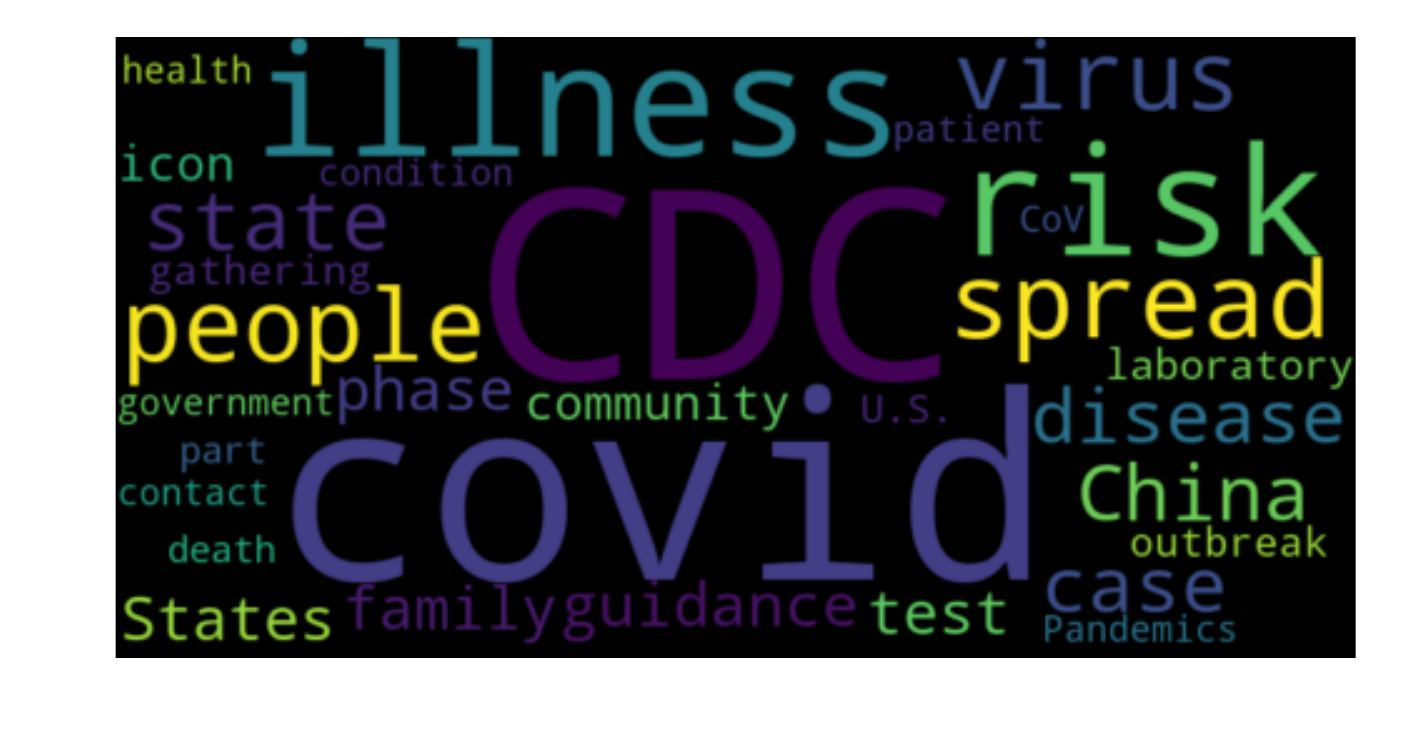} 
\FIG{qcovid}{Word-cloud of the CDC COVID-19 document after query-driven personalized PageRank}{0.72}{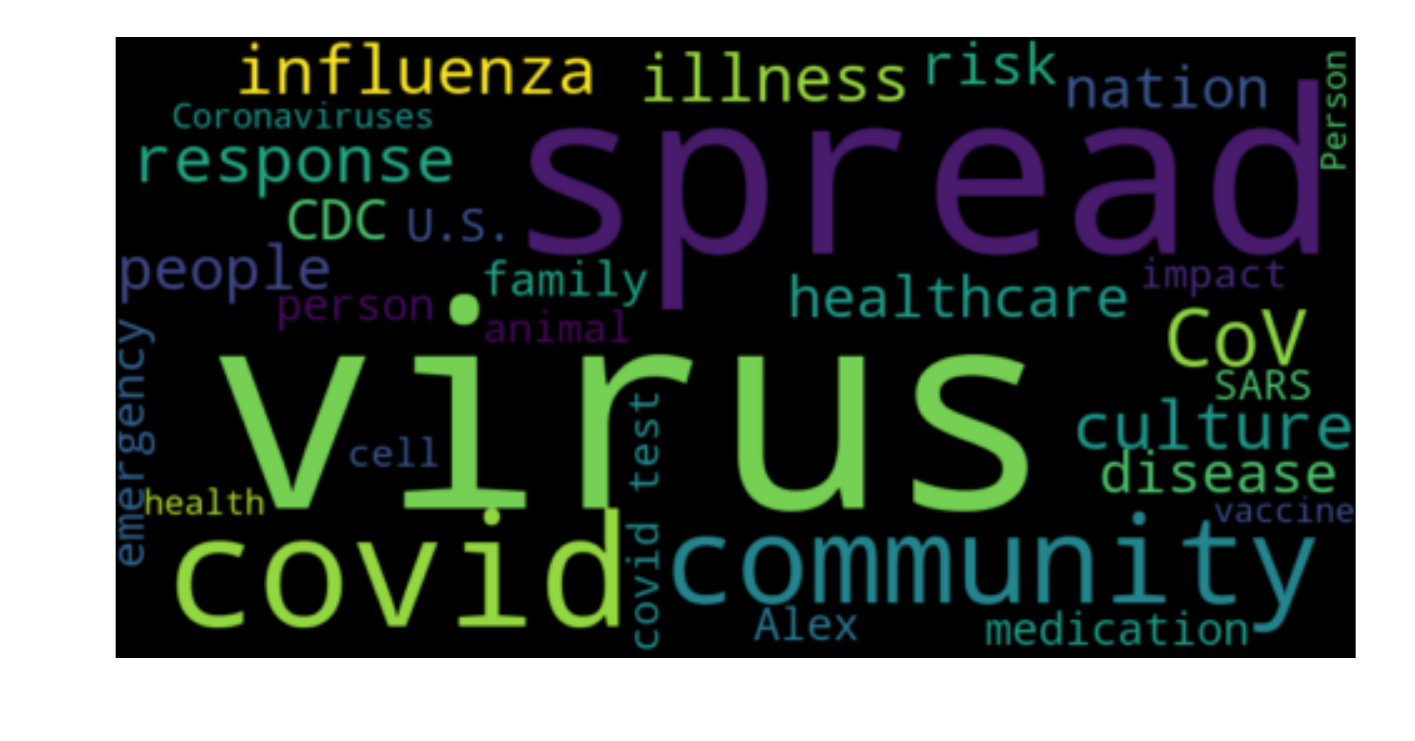}

The following examples show the effect if the {\bf wh-}words triggering
answers derived from the {\tt ner/2} facts.
{
\begin{quote}
{\verb~>>>~ talk\_about('examples/covid')}

{\verb~?--~ {\bf Who did  declare the outbreak an emergency?}}

On March 13 , the President of the United States declared the COVID-19 outbreak a national emergency . 

27 : On March 11 , the COVID-19 outbreak was characterized as a pandemic by the WHO . 
\end{quote}
}

{
\begin{quote}
{\verb~?--~ {\bf When did the outbreak become a pandemic?}}

4 : On March 11 , WHO publicly characterized COVID-19 as a pandemic . 

22 : COVID-19 Now a Pandemic A pandemic is a global outbreak of disease . 

27 : On March 11 , the COVID-19 outbreak was characterized as a pandemic by the WHO .
\end{quote}
}

\section{Discussion}\label{disc}

Ideally, one would like to evaluate the quality of natural language understanding of an AI system by querying it not only about a set of relations explicitly extracted in the text, but also about relations inferred from the text. Moreover, one would also like to have the system justify the inferred relations in the form of a proof, or at least a sketch of the thought process a human would use for the same purpose.
The main challenge here is not only that theorem-proving logic is hard, (with first-order classical predicate calculus already Turing-complete), but also that modalities, beliefs, sentiments, hypothetical and counterfactual judgments often make the underlying knowledge structure intractable.

On the other hand, simple relations, stated or implied by text elements that can be mined or inferred from a ranked graph built from labeled dependency links, provide a limited but manageable approximation of the text's deeper logic structure, especially when aggregated with generalizations and similarities  provided by WordNet or the much richer Wikipedia knowledge graph. 

Given its effectiveness as an interactive content exploration tool, we plan future work on packaging our dialog engine as a set of Amazon Alexa skills for some popular Wikipedia entries as well as  product reviews, FAQs and user manuals.

Empirical evaluation of our keyphrase and summarization algorithms will be subject to a different paper, but preliminary tests indicate that both of them match or exceed Rouge scores for state of the art systems \cite{deep19}.

\section{Related work}\label{rel}
\subsubsection*{Dependency parsing}

The Stanford neural network based dependency parser  \cite{stan} is now part of the Stanford CoreNLP toolkit,\footnote{\url{https://stanfordnlp.github.io/CoreNLP/}} which also comes with  part of speech tagging, named entity recognition and co-reference resolution \cite{coreNLP}.
Its evolution toward the use of Universal Dependencies \cite{ud14} makes
systems relying on it potentially portable to
over {\bf 70}  languages  covered by the Universal Dependencies effort.
\footnote{\url{https://universaldependencies.org/}}
 
Of particular interest is the connection of dependency graphs  to logic elements like predicate argument relations \cite{Choi:2011}.
The automatic conversion of constituency trees to dependency graphs proposed by \citeN{choi:17a} provides a bridge allowing the output of high-quality statistically trained phrase structure parsers to be reused for extraction of dependency links.

{\em  In this context, our novel contribution is that we analyze
dependency links and part-of-speech tags associated to their endpoints
in order to  build a unified document graph from which we
extract SVO relations. By redirecting links to focus
on nouns and sentences we not only enable keyphrase and summary
extraction from the resulting document graph
but also facilitate its use for query answering 
in our dialog engine.
}

\subsubsection*{Graph based Natural Language Processing}

TextRank \cite{EMNLP:TR,ijcnlp05} extracts keyphrases 
using word co-occurrence relations controlled by the distance between words:
two vertices are connected if their corresponding lexical units co-occur within a sliding window ranging from 2 to 10 words.
Sentence similarity is computed as content overlap giving weights to the links that refine the original PageRank algorithm \cite{page98pagerank,brin98anatomy}. TextRank needs elimination of stop words and obtains best results when links are restricted to nouns and adjectives.
\citeN{Erkan:2004} explore several graph centrality measures,
and
\citeN{radabook} offer a comprehensive overview of graph-based natural language processing and related graph algorithms.
Graph-based and other text summarization techniques are surveyed by \citeN{NenkovaM12} and more recently by \citeN{textSum}.
Besides ranking, elements like coherence via similarity with previously chosen sentences
and avoidance of redundant rephrasings are shown to contribute to the overall quality of
the summaries.

{\em 
The main novelty of our approach in this context is building text graphs from dependency links and integrating words and sentences in the same text graph, resulting in a unified algorithm that also enables relation extraction and interactive text mining.
}

\subsubsection*{Relation Extraction}

The relevance of dependency graphs for relation extraction has been identified in several papers.
Among others, \citeN{AdolphsXLU11} point out to their role as a generic interface between parsers and relation extraction systems. 
\citeN{depRelPat} identify several  models grounded on syntactic patterns (e.g., subject-verb-object) that can be mined out from dependency graphs.
Of particular interest for relation extraction facilitated by dependency graphs is the shortest path hypothesis that prefers relating entities like predicates and arguments that are connected via a shortest path in the graph \cite{Bunescu:2005}.
To facilitate their practical applications to biomedical texts, 
\citeN{biorel} extend dependency graphs with richer sets of semantic features including ``is-a'' and ``part-of'' relations and co-reference resolution.

The use of ranking algorithms in combination with WordNet synset links for word-sense disambiguation goes back as far as \citeN{coling04:pr}, which is in fact a prequel to
TextRank \cite{EMNLP:TR}.
With the emergence of resources like Wikipedia, a much richer set of links and content elements has been used in connection with graph-based natural language processing  
\cite{wikiRank,AdolphsXLU11,wikify}.

We currently extract our relations directly from the dependency graph
and by using one step up and one step down links in the WordNet hypernym and meronym hierarchies.
We plan extensions to integrate Wikipedia content via the {\tt dbpedia} database,\footnote{\url{https://wiki.dbpedia.org/}} and to extract more elaborate logic relations using a Prolog-based semantic parser like Boxer \cite{boxer15}.

\subsubsection*{Logic Programming Systems for Natural Language Processing}

A common characteristic of Prolog or ASP-based NLP systems is their focus on closed domains with domain-specific logic expressed in clausal form \cite{actlan17,inclezan18,baral19,inclezan19},
although recent work (e.g., \citeN{naraction}) extracts action language programs from more general narratives.

{\em
As our main objective is the building of a practically useful dialog agent, and as we work with open domain text and query driven content retrieval, our focus is not on precise domain-specific reasoning mechanisms. By taking advantage of the Prolog representation of a document's content, we use reasoning about the extracted relations and ranking information to find the most  relevant sentences derived from a given query and the recent dialog history.
}

\section{Conclusions}\label{conc}

The key idea of the paper has evolved from our search for synergies between symbolic AI and emerging natural language processing tools built with machine learning techniques. It is our belief that these are complementary and that by working together they will take significant forward steps in natural language understanding.
We have based our text graph on heterogeneous but syntactically and semantically meaningful text units (words and sentences) resulting in a web of interleaved  links.
These links  mutually recommend each other's highly ranked instances. Our  fact extraction algorithm, in combination with the Prolog interface,  has elevated the
syntactic information provided by dependency graphs with semantic elements
ready to benefit from logic-based inference mechanisms.
Given the standardization brought by the use of {\em Universal Dependencies}, 
our techniques are  likely to be portable to a large number of languages.

The Prolog-based dialog engine supports spoken interaction with a conversational agent that exposes salient content of the document driven by the user's interests. Its applications range from assistive technologies to visually challenged people, voice interaction with user manuals, teaching from K-12 to graduate-level classes, and interactive information retrieval from complex technical or legal documents.

Last but not least, we have used our system's front end to generate
a Prolog dataset
derived from more than 2000 research papers.
We make this dataset available to other researchers using logic programming based reasoners and content mining tools.\footnote{\url{http://www.cse.unt.edu/~tarau/datasets/PrologDeepRankDataset.zip}}

\section*{Acknowledgment}
We are thankful to the anonymous reviewers of {\bf PADL'2020} for their careful reading and constructive suggestions.

\bibliographystyle{acmtrans}
\bibliography{nlp,nlp1,net,theory,tarau,proglang,biblio}

\end{document}